\documentclass[5p,times,procedia]{elsarticle}
\usepackage{ecrc}

\usepackage{soul,color,xcolor}
\soulregister{\cite}7

\volume{00}

\firstpage{1}

\journalname{Procedia CIRP}

\runauth{Z. Li et al.}

\jid{trpro}

\usepackage{amssymb}

\usepackage[figuresright]{rotating}

\usepackage[bookmarks=false]{hyperref}
    \hypersetup{colorlinks,
      linkcolor=blue,
      citecolor=blue,
      urlcolor=blue}

\begin{document}
\begin{frontmatter}



\dochead{57th CIRP Conference on Manufacturing Systems 2024 (CMS 2024)}

\title{Deep Learning Based Tool Wear Estimation Considering Cutting Conditions}


\author[]{Zongshuo Li\corref{*}\textsuperscript{a,*}}
\author[a]{Markus Meurer}
\author[a,b]{Thomas Bergs}

\address[a]{Manufacturing Technology Institute (MTI) of RWTH Aachen University, Campus-Boulevard 30, 52074 Aachen, Germany}
\address[b]{Fraunhofer Institute for Production Technology IPT, Steinbachstr. 17, 52074 Aachen, Germany}

\aucores{* Corresponding author. Tel.: +49-241-80-20522 ; fax: +49-241-80-22293. {\it E-mail address:} z.li@mti.rwth-aachen.de}

\begin{abstract}
Tool wear conditions impact the final quality of the workpiece. In this study, we propose a deep learning approach based on a convolutional neural network that incorporates cutting conditions as extra model inputs, aiming to improve tool wear estimation accuracy and fulfill industrial demands for zero-shot transferability. Through a series of milling experiments under various cutting parameters, we evaluate the model's performance in terms of tool wear estimation accuracy and its transferability to new fixed or variable cutting parameters. The results consistently highlight our approach’s advantage over conventional models that omit cutting conditions, maintaining superior performance irrespective of the stability of the wear development or the limitation of the training dataset. This finding underscores its potential applicability in industrial scenarios.

\end{abstract}


\end{frontmatter}




\section{Introduction}
\label{main}

Tool wear represents an inevitable aspect of the cutting process where the tool is exposed to complex collective loads due to the interplay of multiple physical fields. This interaction leads to the detachment of surface material from the tool, altering its geometry. Such changes significantly influence the quality of the workpiece and the efficiency of the machining process~\cite{Feng2022}. Consequently, tool wear condition monitoring is important for ensuring workpiece quality and productivity within advanced manufacturing systems.

Changes in tool geometry and other characteristics due to tool wear affect the manufacturing process, such as cutting forces, vibration frequencies, etc. On the other hand, recent sensor technology and signal processing advances have facilitated the capture of those changes from processes~\cite{Li2022} and further contributed to the development of machine learning models for tool wear monitoring, which are trained on historical datasets and estimate tool wear from real-time data inputs~\cite{Nasir2021}. Initial explorations employed various conventional machine learning algorithms, including Random Forests~\cite{Cardoz2023,Wu2017}, Support Vector Machines~\cite{Benkedjouh2013}, and Artificial Neural Networks~\cite{Paul2012}. These approaches rely on feature engineering to extract relevant features from raw signals. For instance, Wang et al. extracted 54 features from cutting force and acceleration signals, applying feature dimensionality reduction and Support Vector Regression to estimate current tool wear~\cite{Wang2017}. However, as monitoring data volume increases, the complexity of feature engineering challenging the real-time applicability of such methods.

Deep Learning (DL) provides a comprehensive approach for tool wear monitoring by integrating feature extraction and prediction, overcoming traditional challenges through its ability to learn feature representations automatically~\cite{Li2022_3}. For example, Li et al. employed a Bidirectional Long Short-Term Memory network to extract deep features from time-series signals for tool wear estimation~\cite{Li2022_4}. Convolutional Neural Networks (CNN) are extensively applied to analyze both image and sequence data due to their robust feature extraction capabilities~\cite{Lecun1998}. Yan et al. converted the force and acceleration signals to 2D frequency domain signals for CNN-based residual network analysis, archieving error margins below 8~\%~\cite{Yan2021}. Huang et al. introduced a multi-scale CNN with an attention fusion module for more accurate and effective tool wear classification~\cite{Huang2022}. However, those approaches used randomly sliced test sets, implying that the tools in the test set have been exposed to the models during training, potentially leading to an overestimation of model performance. Such an approach does not accurately reflect the industrial context of tool wear monitoring, where tools are new and not previously encountered in the training set and the wear progression of each tool is individual.

{}
\vspace*{8pt}
\begin{nomenclature}{}{}
\begin{deflist}[AAAAAA]
\defitem{K}\defterm{Number of sensor signal channels}
\defitem{L}\defterm{Sensor signal length}
\defitem{H}\defterm{Number of available cutting parameters}
\defitem{Conv.}\defterm{Convolution}
\defitem{$v_c$}\defterm{Cutting speed}
\defitem{$f_z$}\defterm{Feed per tooth}
\defitem{$\rm M_T$}\defterm{Tool torque from RCD}
\defitem{$\rm F_{RCDx}, F_{RCDy}$}\defterm{Cutting force components from RCD}
\defitem{$\rm F_{SDx}, F_{SDy}$}\defterm{Cutting force components from SD}
\defitem{$\rm F_{RCDxy}, F_{SDxy}$}\defterm{Resultant forces from RCD/SD}
\defitem{$\rm M_S$}\defterm{Spindle torque}
\defitem{$\rm I_S$}\defterm{Spindle current}
\defitem{$\rm M_x, M_y$}\defterm{Torques of the auxiliary drives}
\defitem{$\rm I_x, I_y$}\defterm{Currents of the auxiliary drives}
\defitem{$\rm M_{xy}$}\defterm{Resultant torque of the auxiliary drives}
\defitem{$\rm I_{xy}$}\defterm{Resultant current of the auxiliary drives}
\defitem{$\rm VB_{mean}^n$}\defterm{Mean width of flank wear land, zone n}
\defitem{$\rm VB_{max}^n$}\defterm{Maximum width of flank wear land, zone n}
\defitem{$\rm VB_{mean}$}\defterm{Global mean width of flank wear land}
\defitem{$\rm VB_{max}$}\defterm{Global maximum width of flank wear land}
\defitem{$\rm R^2$}\defterm{Coefficient of determination}
\defitem{$\rm l_f$}\defterm{Feed travel}
\end{deflist}
\end{nomenclature}\vskip24pt

In practical scenarios, the monitoring of tool wear is complicated by the variability of cutting conditions. The primary challenge is that while sensor captures the tool's wear progression, it may not adequately account for the variations caused by differing cutting conditions. Ignoring of cutting conditions can result in a loss of information, undermining the model's ability to accurately estimate tool wear under new cutting conditions. This issue highlights the necessity of integrating cutting conditions as inputs into the model to enhance the precision of wear estimations across diverse cutting conditions. One of the important cutting conditions is the cutting parameters. Cheng et al. mitigated this by using feature normalization to remove the effects of varying cutting parameters~\cite{Cheng2022}. Li et al. employed meta-learning for rapid model fine-tuning with a minimal dataset under new cutting parameters~\cite{Li2019}. Despite these advancements, the requirement for feature engineering and the collection of sample data under new parameters remain. Ideally, a model would be capable of zero-shot transfer, where it can be directly applied to monitor tool wear in production without the need for retraining or additional data under new cutting parameters. However, given the diversity of possible cutting parameters encountered in production, which may not be fully represented in the training data, achieving true zero-shot transferability in an industrial context remains a challenge.

This study aims to introduce a deep learning approach using CNN that incorporates cutting conditions to enhance tool wear estimation accuracy under various cutting parameters. Compared to previous studies, this approach eliminates complex feature engineering and enables the model to adapt to new cutting parameters without retraining, aligning with zero-shot transfer requirements for industrial application. To assess the model's accuracy and transferability, milling experiments were conducted under various cutting parameters. Chapter 2 outlines the structure of the model. Chapter 3 details the experimental setup and data processing flow. Chapter 4 analyzes and discusses the validation results. The final chapter summarizes the characteristics of the proposed approach and provides an outlook on its future development.

\section{Deep Learning Approach Considering Cutting Parameters}

\begin{figure}[h]
\centering\includegraphics[width=0.9\linewidth]{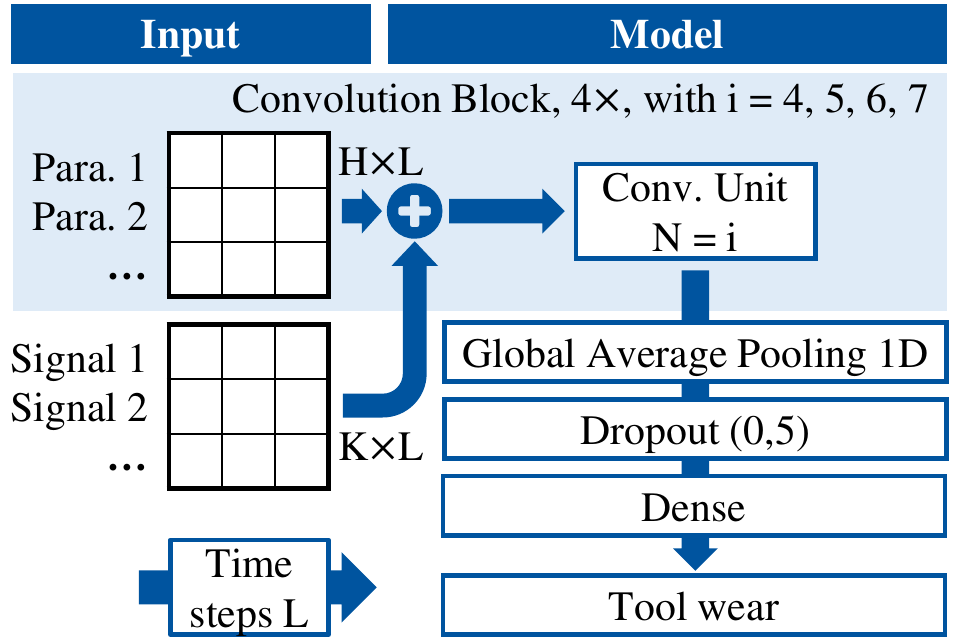}
\caption{Concept of the proposed approach}\label{fig:2_1_CNN}
\end{figure}

The model is based on a CNN designed to estimate tool wear under specific cutting conditions. The architecture of the model is illustrated in Figure~\ref{fig:2_1_CNN}. 


The model gets K time series signals of length L obtained from sensors as input. These signals are formatted into an K$\times$L dimensional signal sequence for processing within the convolution blocks. In the convolution blocks, H available cutting parameters are copied to align with the input time series of length L, generating a parameter sequence of dimensions H$\times$L. This parameter sequence is subsequently combined with the signal sequence, formed in an (K+H)$\times$L dimensional parameter-signal sequence that is fed into the convolution unit. The convolution unit is structured with two one-dimensional convolutions, each employing $\rm 2^N$ filters, followed by a max pooling operation with a pool size of 3. To enhance the prompting of cutting parameters to the model, the output from the convolutional unit is then combined with the parameter sequence of the corresponding length again before entering the next convolutional unit, with this process repeating four times. The final stage of the model consists of global average pooling, dropout, and a dense layer. The model is designed to multiple target values related to tool wear, aiming to enhance both the performance and robustness through the incorporation of multiple detailed tool wear measurements. Detailed descriptions of the input signals and tool wear measurements are deferred to Chapter 3. The primary objective of this study is to demonstrate the enhanced performance of the model when cutting conditions are considered (Test Model), in comparison to a baseline model where such conditions are not considered (Reference Model). We utilized a grid search method to identify the optimal number of layers and the number of filters in the first layer to optimize the network performance. The architecture follows the standard design of a classical convolutional neural network. It is important to note that this paper's scope is limited to this comparative analysis and does not extend to further optimization strategies, such as the incorporation of residual networks or recurrent neural networks (RNN). Both the test and reference models share identical training configurations, ensuring a consistent basis for comparison.

\section{Materials and Methods}
This chapter details the experimental setup and data processing flow.

\subsection{Experimental setup}

\begin{figure}[h]
\centering\includegraphics[width=0.9\linewidth]{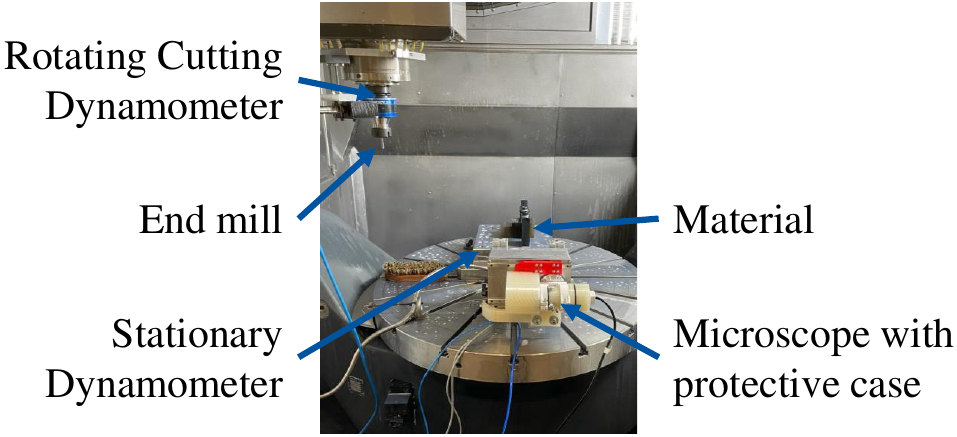}
\caption{Experimental Setup}\label{fig:3_1_Experimental_Setup}
\end{figure}

A total of 20 milling experiments (corresponding to 20 tools) were carried out under different cutting parameters, utilizing a 5-axis machining center DMU 85 Monoblock from DMG Mori. Figure~\ref{fig:3_1_Experimental_Setup} shows the experimental setup.


The workpiece material was Inconel 718 with dimensions of a 50~mm cube. The uncoated end mill featuring a HM MG10 substrate, four cutting edges, a diameter of 6~mm, and a corner radius of 0.15~mm was employed. Each cut was executed along the X-axis of the machine, covering a 50~mm feed travel. All four peripheral cutting edges of the end mill were inspected after each cut using a DinoLite optical microscope, ensuring a comprehensive documentation of wear development. The cutting force components on the workpiece side were measured via a Kistler 9255C Stationary Dynamometer (SD), while the tool-side torque and cutting force components were measured using a Kistler 9170A Rotary Cutting Dynamometer (RCD), both at a 10 kHz sample rate. Additionally, the machine internal signals including spindle torque, spindle current, and auxiliary drive torque and current were recorded synchronously at a lower sample rate of 100 Hz. The depth and width of cut remained constant at 1.5~mm in all experiments. The first 16 sets of experiments were each conducted under consistent cutting parameters, with details provided in Table~\ref{tab:parameters_16}. The last 4 sets introduced variations in cutting parameters to further assess the model's adaptability and transferability, where each cut uses a different parameter. The corresponding tool numbers and sequences of cutting parameters applied are given in Table~\ref{tab:parameters_4}. This comprehensive experimental design aimed to validate the proposed deep learning approach's effectiveness and robustness in estimating tool wear across a spectrum of cutting parameters. 

\begin{table}[h]
\caption{Cutting parameters for the first 16 sets of experiments}\label{tab:parameters_16}
\begin{tabular*}{\hsize}{@{\extracolsep{\fill}}llll@{}}
\toprule
Tool No. & $v_c$ / m/min & $f_z$ / mm & Parameter Set No.\\
\colrule
1, 2   &   30  &  0.03 & 1\\
3, 4   &   40  &  0.04 & 2\\
5, 6   &   20  &  0.03 & 3\\
7, 8   &   20  &  0.04 & 4\\
9, 10  &   30  &  0.02 & 5\\
11, 12 &   30  &  0.04 & 6\\
13, 14 &   40  &  0.02 & 7\\
15, 16 &   40  &  0.03 & 8\\
\botrule
\end{tabular*}
\end{table}

\begin{table}[h]
\caption{Cutting parameters for the last 4 sets of experiments}\label{tab:parameters_4}
\begin{tabular*}{\hsize}{@{\extracolsep{\fill}}llll@{}}
\toprule
Tool No. & Parameter Set Sequence\\
\colrule
17 & 6 3 5 8 1 7 6 3 8 7 2 \\
18 & 1 4 8 3 3 3 7 1 8 \\
19 & 8 4 3 8 6 1 1 5 8 7 6 4 1 5 \\
20 & 3 8 6 5 4 6 1 8 8 4 6 2 \\
\botrule
\end{tabular*}
\end{table}

\subsection{Data preparation}
Data acquired via sensors and microscopes is subjected to preprocessing to ensure its suitability for modeling. Specifically, in a milling operation, which encompasses three distinct phases: cut-in, milling, and cut-out. The milling phase is characterized by uniform cutting conditions. Consequently, signal segments pertinent to the milling phase are isolated based on the auxiliary drive position, with only the segments from the last two seconds being selected for model input. The experimental setup delineates signal sources into two categories: external sensor signals and machine internal signals. External sensor signals are equipped with seven signal channels, including tool torque measured by RCD $\rm M_T$, two orthogonal cutting force components from RCD ($\rm F_{RCDx}$ and $\rm F_{RCDy}$), and two orthogonal cutting force components from SD ($\rm F_{SDx}$ and $\rm F_{SDy}$). The resultant forces, $\rm F_{RCDxy}$ or $\rm F_{SDxy}$, can be computed from the respective orthogonal components. On the other hand, machine internal signals comprise eight signal channels, including the spindle torque $\rm M_S$ and the spindle current $\rm I_S$, along with the orthogonal torques ($\rm M_x$ and $\rm M_y$) and currents ($\rm I_S$ and $\rm I_S$) of the auxiliary drive. The resultant torque $\rm M_{xy}$ and current $\rm I_{xy}$ for these signals are also calculated. Given the external sensors' sample rate of 10 kHz, machine internal signals are subjected to linear interpolation to align with the external sensor signal length. Consequently, for each milling operation, the processed external sensor signals results in a signal sequence dimension of 7$\times$20,000, while machine internal signals yield a dimension of 8$\times$20,000. Table~\ref{tab:signal_sets} gives an overview of the signals.

\begin{table}[ht]
\caption{Overview of all signals}\label{tab:signal_sets}
\begin{tabular*}{\hsize}{@{\extracolsep{\fill}}lll@{}}
\toprule
External sensor signals & Machine internal signals\\
\colrule
$\rm M_T$                                         & $\rm M_S$, $\rm I_S$ \\
$\rm F_{RCDx}$, $\rm F_{RCDy}$, $\rm F_{RCDxy}$   & $\rm M_x$, $\rm M_y$, $\rm M_{xy}$ \\
$\rm F_{SDx}$, $\rm F_{SDx}$, $\rm F_{SDxy}$      & $\rm I_x$, $\rm I_y$, $\rm I_{xy}$ \\
\botrule
\end{tabular*}
\end{table}

\begin{figure}[h]
\centering\includegraphics[width=0.9\linewidth]{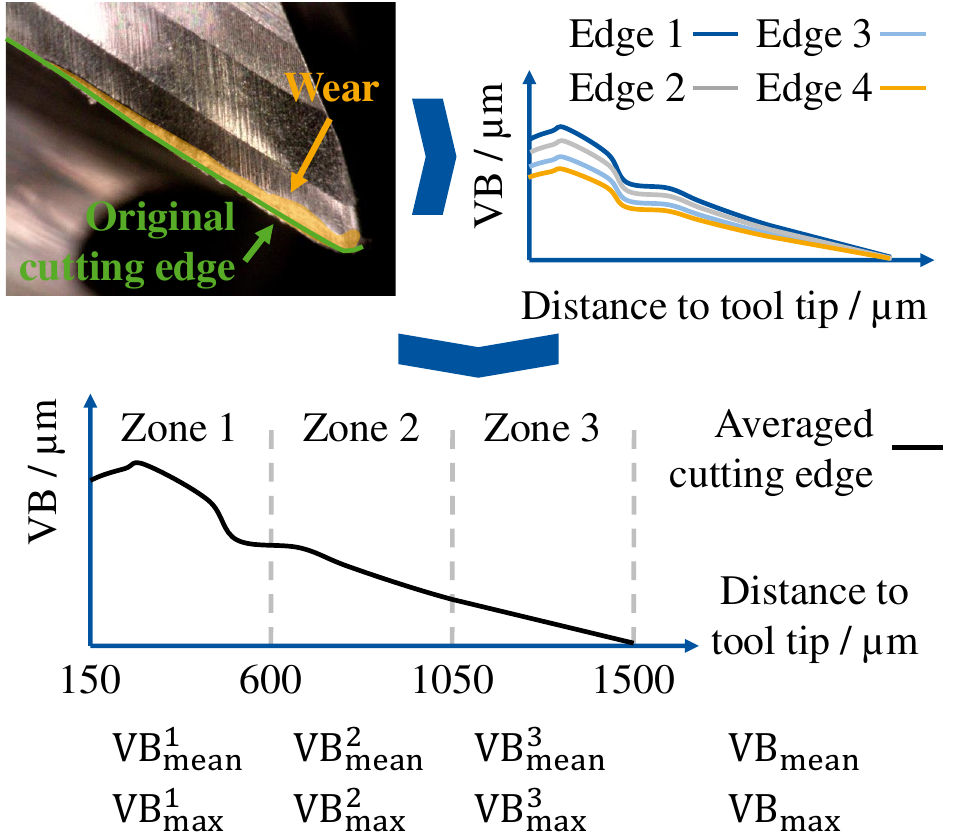}
\caption{Measurement process}\label{fig:3_2_Measurement}
\end{figure}

The model's output comprises eight distinct measurements of flank wear land width, a process illustrated in Figure~\ref{fig:3_2_Measurement}. The process initiates with the conversion of the widths of flank wear land from four cutting edges of one end mill into four measurement curves. These curves represent the widths of flank wear land depending on the distance from the tool tip. Subsequently, these four curves are averaged to produce a singular average curve. This average curve is then segmented into three distinct zones, each spanning 450~µm, to represent three zones at varying distances from the tool tip. For each of these zones, both the mean and maximum widths of flank wear land are calculated, denoted as $\rm VB_{mean}^n$ and $\rm VB_{max}^n$, where n represents the specific zone. Furthermore, an analysis of the entire curve yields global mean and maximum width of flank wear land, denoted as $\rm VB_{mean}$ and $\rm VB_{max}$. By deriving eight widths of flank wear land from a single measurement, the model gains a broader array of learning targets, thereby enhancing its overall performance and robustness.


\section{Results and Discussion}

In the experiment, a series of 454 milling operations followed by wear measurements were conducted with 20 tools. The global maximum width of flank wear land $\rm VB_{max}$ was selected for evaluation, as it serves as a key indicator for tool replacement. Model performance was assessed using two primary metrics: the Root Mean Square Error (RMSE) and the Coefficient of Determination ($\rm R^2$). The model's zero-shot transferability was examined under two distinct scenarios. One where the test sets comprised experiments with fixed cutting parameters (the first 16 experiments) and another where the test set involved experiments with variable cutting parameters (the last 4 experiments). For comparative analysis, a reference model with identical model architecture but not incorporating cutting parameters was employed as a baseline. This comparison facilitated a direct assessment of the impact of incorporating cutting parameters on the model's performance and its transferability.

\subsection{Transfer to a fixed cutting parameter}


\begin{figure}[b]
\centering\includegraphics[width=0.9\linewidth]{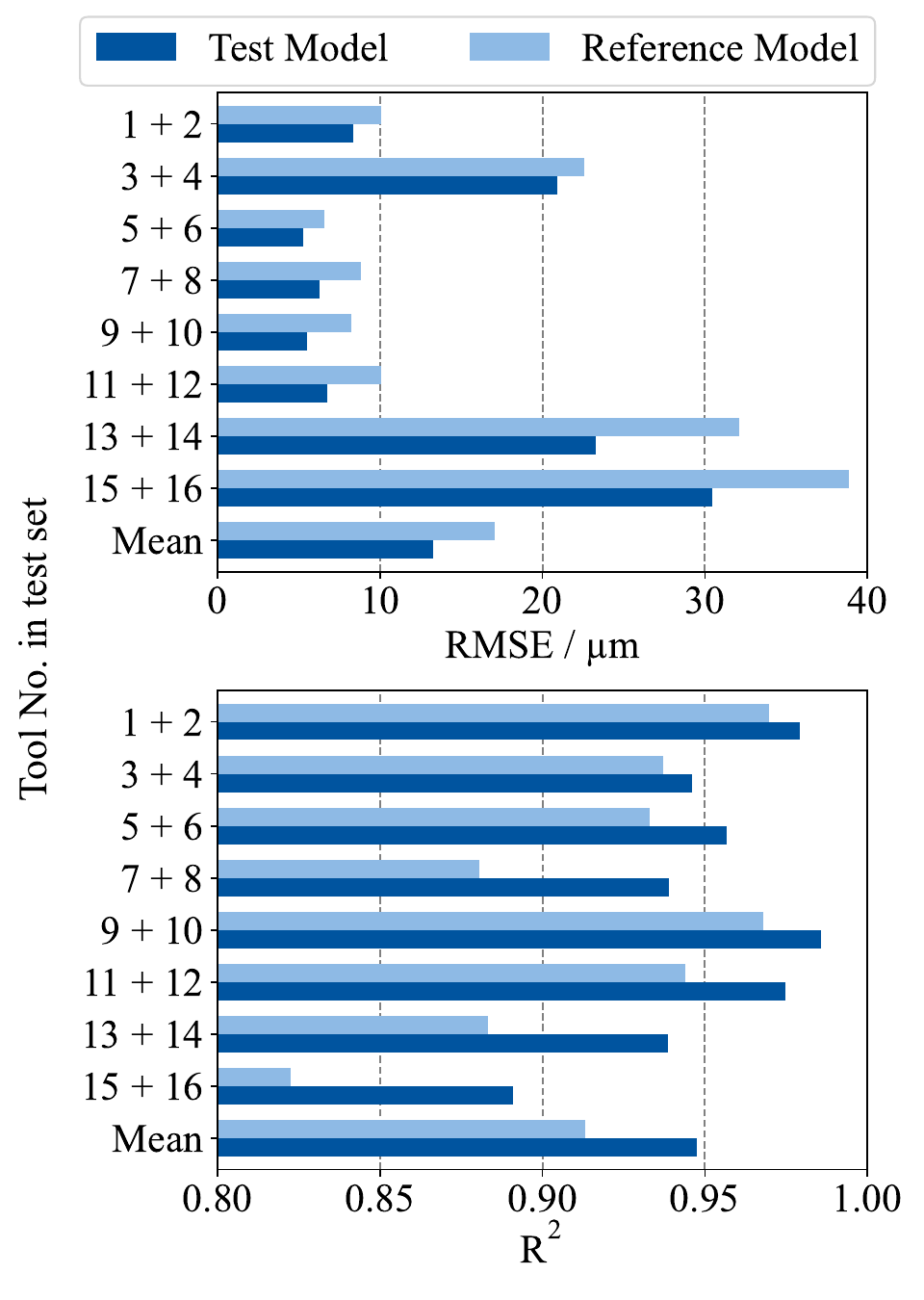}
\caption{Comparative transferability to a fixed cutting parameter}\label{fig:4_1_Barh_Phase1_RMSE_R2}
\end{figure}
 

In this evaluation scenario, the experimental data were grouped into eight sets based on different cutting parameters. Each set was sequentially designated as the test set, while the remaining sets served as the training set. The model uses all external sensor signals and machine internal signals as inputs.

Figure~\ref{fig:4_1_Barh_Phase1_RMSE_R2} illustrates the comparative performance in tool wear estimation between the test model and a reference model across various test sets. The inclusion of cutting parameters significantly enhances the test model's performance, yielding a 22.3~\% improvement in RMSE and a 3.8~\% increase in the coefficient of determination $\rm R_2$ on average over the reference model. The improvement difference between RMSE and $\rm R^2$ indicates that the reference model is subject to a higher incidence of large estimation errors, leading to a higher RMSE value. Conversely, the test model demonstrates a more uniform error distribution. It is important to note that both models exhibit an RMSE exceeding 20~µm at a cutting speed $v_c$ of \mbox{40 m/min}, yet the test model maintains an 18.3~\% advantage. A possible explanation for this diminished performance at higher speeds is the significant rise in cutting edge temperature, which softens the substrate and accelerates wear, thus destabilizing the wear development. Moreover, the constant cutting fluid flow rate becomes less effective at higher speeds due to reduced cutting fluid per tool engagement, deteriorating the tool's cooling and lubrication conditions. This situation escalates the wear development, introduces signal noise, and thereby complicating the feature extraction process for the model. Particularly, when employing tools No. 3 and No. 4 as the test set, the test model's RMSE advantage drops to 7.5~\%. However, at a lower cutting speed $v_c$ of 20 or \mbox{30 m/min}, where wear progression is more stable, the test model's performance significantly improves. For instance, with tools No. 9 and No. 10 as the test set, the test model achieves an RMSE of merely 5.5~µm, which represents a 33.2~\% improvement over the reference model. These observations emphasize the efficacy of our approach in identifying the relationship between cutting parameters and signal features, enabling robust zero-shot transferability. Crucially, our approach consistently outperforms the reference model, irrespective of the wear development's stability, highlighting its reliability and effectiveness in tool wear estimation.

\subsection{Transfer to variable cutting parameters}
In this evaluation scenario, the experimental data of the initial 16 experiments were utilized as the training dataset, while the data of the final four experiments with tool No. 17 to No. 20 constituted the test set. Diverging from the fixed cutting parameters applied in the training set, each tool in the test set was subjected to varying cutting parameters during each cut. Given that the training set contains all eight cutting parameters, the model is theoretically equipped to identify sensor data features in this scenario. However, the development of tool wear in the test set diverged from the uniform progression observed in the training set, reflecting more realistic industrial applications where tools are subjected to varying cutting parameters throughout their lifecycle. The model leverages both external sensor signals and machine internal signals as input. 

\begin{figure}[t]
\centering\includegraphics[width=0.9\linewidth]{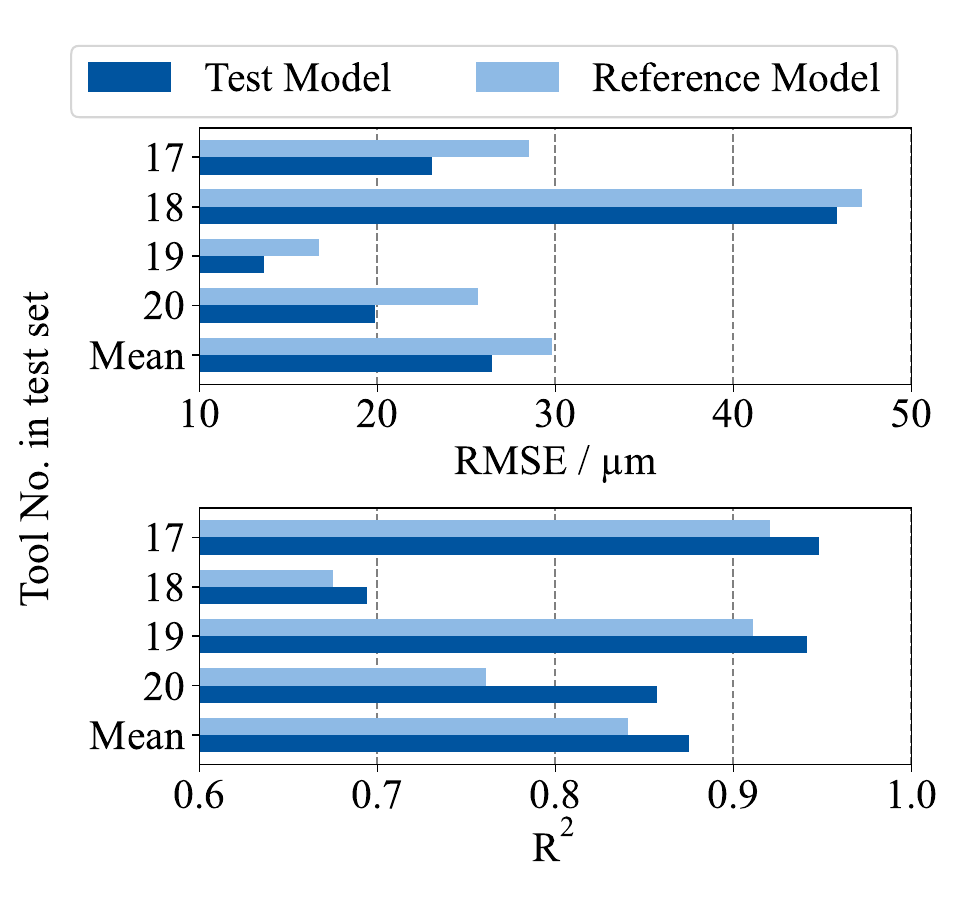}
\caption{Comparative transferability to variable cutting parameters}\label{fig:4_2_Barh_Phase2_RMSE_R2}
\end{figure}

Figure~\ref{fig:4_2_Barh_Phase2_RMSE_R2} illustrates the wear estimation performance of both the test and reference models for the four tools within the test set. The test model demonstrates an average performance improvement over the reference model, with an 11.4~\% reduction in RMSE and a 4.1~\% increase in $\rm R^2$. Despite this advantage, the average RMSE for both models exceeds 25~µm, with the reference model peaking at an RMSE of 47.3~µm. The comparison to the first evaluation scenario indicates a decline in performance for both models, and a narrowing in the performance gap, as evidenced by the reduced advantage in RMSE for the test model. This outcome indicates that the test model, despite outperforming the reference model, still encounters large wear estimation errors.


\begin{figure}[b]
\centering\includegraphics[width=0.9\linewidth]{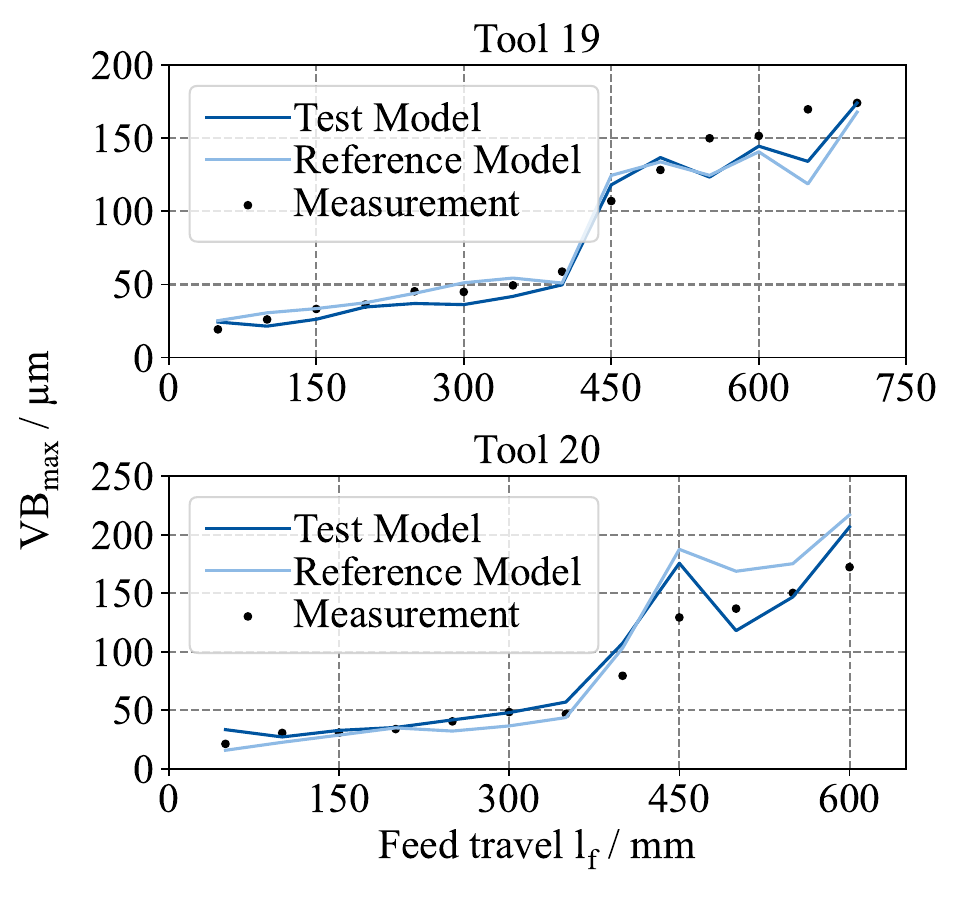}
\caption{Measured and estimated wear curves for tools No. 19 and No. 20}\label{fig:4_3_Plot_Phase2_T2122}
\end{figure}


Figure~\ref{fig:4_3_Plot_Phase2_T2122} presents the measured and estimated wear curves for tools No. 19 and No. 20. Initially, a cutting speed of 20 or \mbox{30 m/min} was applied, leading to a steady wear progression. For tool No. 19, a cutting speed of \mbox{40 m/min} was utilized between feed travel of 400 to 500~mm, and similarly for tool No. 20, the same cutting speed was applied from 350 to 450~mm feed travel. This resulted in a sharp increase in wear over this 100~mm interval. Subsequently, the cutting speed with stable wear development was used again. It can be observed that the distribution of the estimation errors of both models is not homogeneous. The large errors start to appear during rapid wear growth phases and persists until the end of tool life. Specifically, at a feed travel of 650~mm for tool No. 19, wear was underestimated by the reference model, exhibiting errors exceeding 50~µm. Similarly, for tool No. 20, beyond a feed travel of 350~mm, the reference model consistently overestimated wear, with a maximum error surpassing 50~µm. This decline in performance can be attributed to several factors. The employment of a cutting speed of \mbox{40 m/min} with unstable wear development complicated feature extraction. Additionally, despite reverting to a cutting speed of 20 or \mbox{30 m/min} later, the absence of sufficient high-wear condition training data under this parameter led to a reduced wear estimation capability. Nonetheless, the test model, while also exhibiting increased error during these phases, maintained a maximum error within 30~µm, demonstrating a relative improvement. Notably, in the latter stages, both the error and the fluctuations were consistently lower compared to the reference model. These findings indicate that despite the training data limitations, our approach maintains a superior performance over the reference model.

\section{Conclusion and Outlook}
Changes in cutting conditions during machining lead to variations in process monitoring data distribution, presenting a challenge for accurate tool wear estimation under new conditions. Addressing the demand for zero-shot transferability within industrial settings, this study proposes a deep learning approach based on CNN that incorporates cutting conditions as extra model inputs. The model’s performance is assessed through milling experiments conducted under diverse cutting parameters. In comparison to models that omit cutting conditions, our model exhibits superior wear estimation accuracy and enhanced zero-shot transferability. This advantage is maintained irrespective of the stability of the wear development or the limitation of the training dataset, underscoring the effectiveness of incorporating cutting conditions as model inputs to refine wear estimation. This effectiveness enables the model to achieve zero-shot transfer. In practical applications, a model trained on data from a limited set of cutting parameters can be directly used to monitor tool wear in manufacturing processes without retraining under new cutting parameters. Additionally, the estimated wear values help identify potential geometric errors and facilitate necessary compensations using adaptive methods~\cite{Eger2020}, ensuring the quality of the final product.

Future work will aim to broaden the cutting parameter space, thereby extending the model's transferability assessment. Additionally, efforts will be directed towards optimizing the model's architecture to further elevate estimation performance. Lastly, the development of an adaptive design of experiments algorithm will be explored, intending to enhance zero-shot transferability with minimal training data, thereby streamlining the model's applicability in diverse machining contexts.

\section*{Acknowledgements}

The authors appreciate the funding of this work by the Deutsche Forschungsgemeinschaft (DFG) – German Research Foundation for the project 509813741 “Entwicklung und Erforschung eines lernfähigen Systems zur Werkzeugverschleißüberwachung auf Basis künstlicher Intelligenz”.

\vfill\pagebreak





\bibliographystyle{elsarticle-num}
\bibliography{PROCIRP_CMS_2024_Reference}












\end{document}